\documentclass[a4paper,twoside]{article}

\usepackage{epsfig}
\usepackage{subcaption}
\usepackage{calc}
\usepackage{amssymb}
\usepackage{amstext}
\usepackage{amsmath}
\usepackage{amsthm}
\usepackage{multicol}
\usepackage{pslatex}
\usepackage{apalike}
\usepackage[bottom]{footmisc}
\usepackage[nameinlink,noabbrev]{cleveref}
\usepackage[export]{adjustbox}

\usepackage{color} 

\usepackage[dvipsnames]{xcolor, colortbl}
\newcolumntype{a}{>{\columncolor{gray!20}}c}

\usepackage{SCITEPRESS}     

\begin{document}

\title{A-Eye: Driving with the Eyes of AI for Corner Case Generation}

\author{\authorname{Kamil Kowol\sup{1} Stefan Bracke\sup{2} and Hanno Gottschalk\sup{1}}
\affiliation{\sup{1}School of Mathematics and Natural Sciences, IZMD, University of Wuppertal, Gaußstraße 20, Wuppertal, Germany}
\affiliation{\sup{2}Chair of Reliability Engineering and Risk Analytics, University of Wuppertal, Gaußstraße 20, Wuppertal, Germany}
\email{\{kowol, bracke, hgottsch\}@uni-wuppertal.de}
}


\abstract{The overall goal of this work is to enrich training data for automated driving with so-called corner cases in a relatively short period of time. In road traffic, corner cases are critical, rare and unusual situations that challenge the perception by AI algorithms. For this purpose, we present the design of a test rig to generate synthetic corner cases using a human-in-the-loop approach. For the test rig, a real-time semantic segmentation network is trained and integrated into the driving simulation software CARLA in such a way that a human can drive on the network's prediction. In addition, a second person gets to see the same scene from the original CARLA output and is supposed to intervene with the help of a second control unit as soon as the semantic driver shows dangerous driving behavior. Interventions potentially indicate poor recognition of a critical scene by the segmentation network and then represent a corner case. In our experiments, we show that targeted and accelerated enrichment of training data with corner cases leads to improvements in pedestrian detection in safety-relevant episodes in road traffic.}

\onecolumn \maketitle \normalsize \setcounter{footnote}{0} \vfill

\section{\uppercase{Introduction}}
\label{sec:introduction}
Despite AI systems achieve impressive performance in solving specific tasks, e.g. in automated driving, they lack understanding of the context of safety in traffic. In contrast, while humans are often described as lousy drivers, as they tend to be diverted or feel fatigue, humans have a fine understanding, when a traffic scene could lead to a situation, where humans are at risk.

It has been observed previously, that to increase robustness and performance of AI algorithms a large number of clean and diverse scenes is needed~\cite{Karpathy2021online}. However, a large amount of annotated data per se might not imply safe operation in those rare situations, where road users are exposed to a substantial risk. In this work, we aim to present an accelerated testing strategy that leverages human risk perception to capture corner cases and thereby achieve performance improvement in safety-critical scenes. In order to obtain many safety-critical corner cases in a short time, we stop training at an early stage so that the network is sufficiently well trained. Nevertheless, the scenes generated in this way are still useful to improve fully trained networks.

For this purpose, a semantic network is trained with synthetic images from the open-source driving simulation software CARLA~\cite{dosovitskiy2017carla}. In addition, a test rig consisting of 2 control units is connected to CARLA in such a way that the ego vehicle can be controlled with both control units by a human. In this process, the semantic segmentation network is integrated into CARLA in such a way that first the original CARLA image is sent through the network and the prediction is displayed on the screen of one driver. The second driver, in turn, sees the real CARLA image and is supposed to intervene as a safety driver only if he or she feels that a situation is being wrongly assessed by the other driver. We aim to consider situations in which the AI algorithms lead to incorrect evaluations of the scene, which we refer to as safety-relevant corner cases, in order to improve performance through targeted data enrichment. This is done by exchanging images from the original dataset with the safety-critical corner cases, thus keeping the total amount of data fixed. We show that the semantic segmentation network that contains safety-critical corner cases in the training data performs better on similar critical situations than the network that does not contain any safety-critical situations.

\begin{figure*}[thpb]
\begin{minipage}{.5\textwidth}
  \includegraphics[width=1\textwidth, frame]{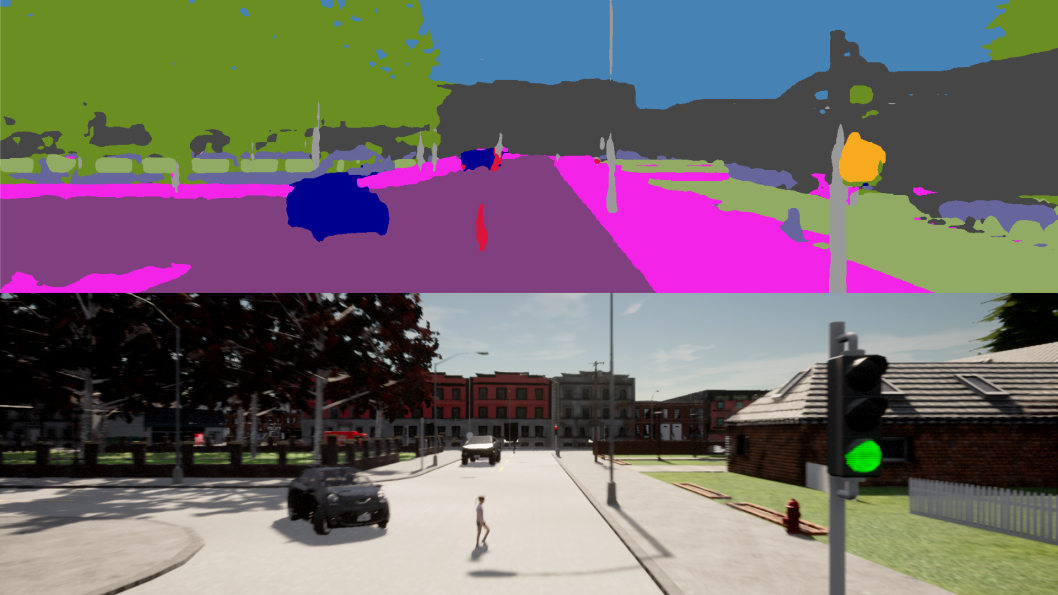}
  \caption{View of the semantic driver (top) and the safety \\driver (bottom).}
  \label{fig:carla_view}
\end{minipage}
\begin{minipage}{.5\textwidth}
  \includegraphics[width=1\textwidth, frame]{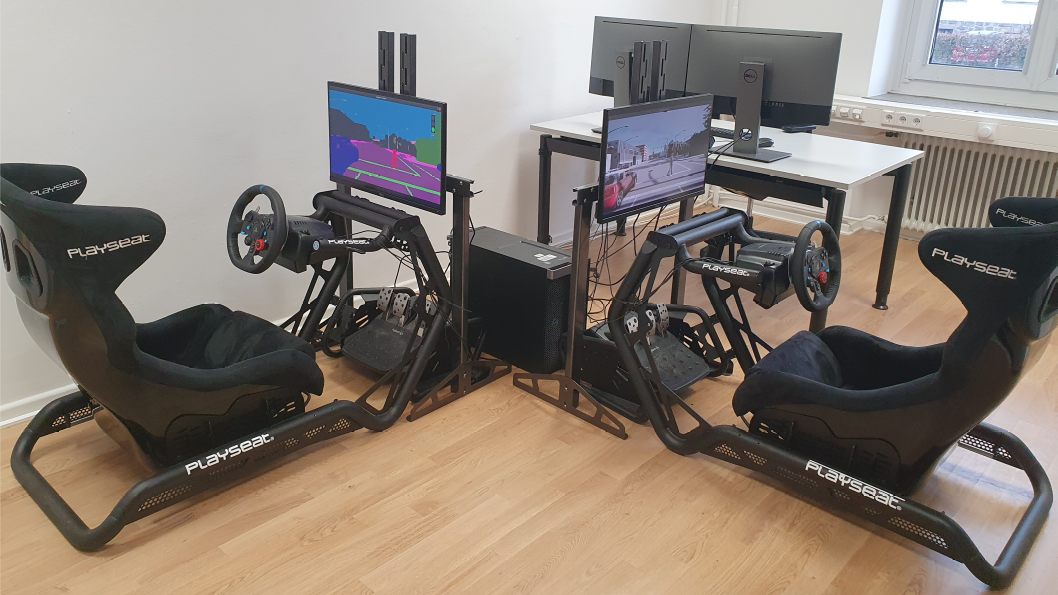}
  \caption{Test rig including steering wheels, pedals, seats and screens.}
  \label{fig:test_rig}
\end{minipage}
\end{figure*}

Our approach somehow follows the idea of active learning, where we get feedback on the quality of the prediction by interactively querying the scene. However, unlike in standard active learning we do not leave the query strategy to the learning algorithm, but make use of the human's fine tuned sense of risk to query safety-relevant scenes from a large amount of street scenes, leading to enhanced performance in safety-critical situations.

\begin{figure*}[t]
\centering
\includegraphics[width=\linewidth, frame, trim={0 0 0 0}, clip]{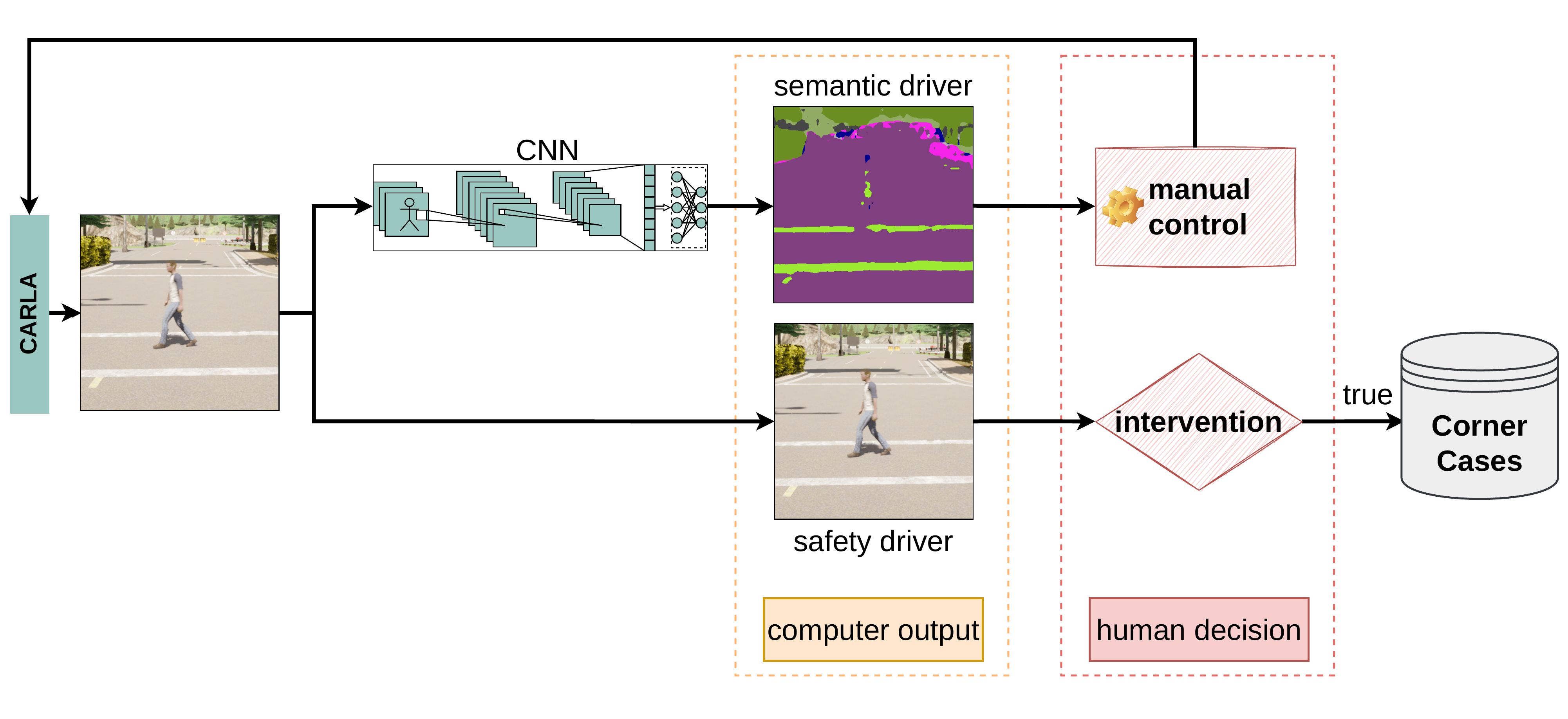}
\caption{Two human subjects can control the ego vehicle. The semantic driver moves the vehicle in compliance with traffic rules in the virtual world and sees only the output of the semantic segmentation network. The safety driver, who sees only the original image, assumes the role of a driving instructor and intervenes in the situation by braking or changing the steering angle as soon as a hazardous situation occurs. Intervening in the current situation indicates poor situation recognition of the segmentation network and represents a corner case. Triggering a corner case ends the acquisition process and a new run can be started.}
\label{fig:flowchart}
\end{figure*}

The contributions of this work can be summarized as follows:
\begin{itemize}
    \item An experimental setup, that could also be implemented in the real world, permits testing the safety of the AI perception separately from the full system safety including the driving policy of an automated vehicle.  
    \item A proof of concept for the retrieval of training data for automated driving with a human-in-the-loop approach that is safety-relevant.
    \item A proof that training on safety-relevant situations generated during poor network performance is beneficial for the recognition of street hazards.
\end{itemize}

\paragraph*{Outline}
\Cref{sec:related_works} discusses related work on corner cases, human-centered AI and human-in-the-loop. In \Cref{sec:setup} we briefly describe the experimental setup used for corner case generation and which data and network were used for our experiments. In \Cref{sec:data_collection} we explain our strategy to generate corner cases. In \Cref{sec:eval} we demonstrate the beneficial effects of training with corner cases for safety-critical situations in automated driving. Finally, we present our conclusions and give an outlook on future directions of research in \Cref{sec:conclusion}.

\section{\uppercase{Related Works}} \label{sec:related_works}
\subsection{Corner Cases}
Training data contains few, if any, critical, rare or unusual scenes, so-called corner or edge cases. In the technical fields, the term corner case describes special situations that occur outside the normal operating parameters~\cite{Chipengo2018cc}. 

According to~\cite{Bolte2019}, a corner case in the field of autonomous driving describes a "non-predictable relevant object/class in relevant location". Based on this definition, a corner case detection framework was presented to calculate a corner case score based on video sequences. The authors of~\cite{breitenstein2020cc} subsequently developed a systematization of corner cases, in which they divide corner cases into different levels and according to the degree of complexity. In addition, examples were given for each corner case level. This was also the basis for a subsequent publication with additional examples~\cite{Breitenstein2021}. Since the approach in these references is camera-based, a categorization of corner cases at sensor level was adapted by~\cite{Heidecker2021}, where RADAR and LiDAR sensors were also considered. Furthermore, this reference presents a toolchain for data generation and processing for corner case detection. 

Outside normal parameters also includes terms such as anomalies, novelties, or outliers, which, according to~\cite{Heidecker2021}, correlate strongly with the term corner case. In road traffic, the detection of new and unknown objects, anomalies or obstacles, which must also be evaluated as 'outside the operating parameters', is essential. To measure the performance of methods for detecting such objects, the benchmark suite "SegmentMeIfYouCan" was created~\cite{chan2021segmentmeifyoucan,chan2021entropy}. In addition, the authors present two datasets for anomaly and obstacle segmentation to help autonomous vehicles better assess safety-critical situations. 

In summary, the term "corner case" can encompass rare and unusual situations that may include anomalies, unknown objects or outliers which are outside of operating parameters. Outside the operating parameters, in the context of machine learning, means that these situations or objects were not part of the training data.

\subsection{Human-Centered AI}
According to the Defense Advanced Research Project Agency (DARPA), the development of AI systems is divided into 3 waves~\cite{DARPAonline,Highnam2020DARPA}. While in the first wave patterns were recognized by humans and linked into logical relationships (crafted knowledge), statistical learning could be used in the second wave due to improved computing power and increased memory capacity. 

We are currently in the third wave, which relates to the explainability and contextual understanding of AI. Here, the black-box approaches, which emerged in the second wave, are to be understood so that AI decision-making becomes comprehensible to humans. This last point is therefore also an important one in the HCAI initiative, which aims to connect different domains with human-centered AI~\cite{hai2018stanford}. 

On this basis, the authors of~\cite{xu2019article,xu2022} developed an extended HCAI framework that defines the following three different design goals for the human-centered AI: \textit{Ethically Aligned Design} that avoid biases or unfairness of AI algorithms such that these algorithms make decisions according to human criteria and rules. \textit{Technology Design} which considers human and machine intelligence to exploit synergies. \textit{Human Factors Design} to make AI-solutions explainable. The aim is to give humans an insight into the decision-making process of AI algorithms, so that trust in the current technology can be increased. For this purpose, we have built a specially developed test rig to incorporate human behavior into the further development of AI in the field of autonomous driving. 
At the same time, we would like to use the test rig to provide a demonstration object to illustrate AI algorithms to society. The focus will be on allowing humans to visually perceive and interact with the decision-making of AI algorithms. In the context of autonomous driving this would mean: driving with the eyes of AI.

\begin{table*}[ht]
\caption{Performance measurement on two test datasets. The comparison shows that the addition of safety-critical scenes in training also improves performance in testing with safety-critical scenes.}
\begin{center}
\scalebox{.8}{
\begin{tabular}{|c|c|c||c|c|c|c|c|}
    \hline
    \multicolumn{3}{|c||}{\textbf{traindata}} & \multicolumn{2}{c|}{\textbf{safety-critical testdata}}& \multicolumn{2}{c|}{\textbf{natural distributed testdata}}\\
    \hline
    \textit{no.}        &       \textit{name} & \textbf{$mean_{pixels/scene}$} & \textbf{$IoU_{pedestrians}$} & \textbf{$mIoU$} &        \textbf{$IoU_{pedestrians}$} & \textbf{$mIoU$}\\
    \hline
    $1$ & natural distribution & $3583.6$   &  $0.4600$             & $0.6954$  &  $0.4937$    & $\textbf{0.761}$\\
    \hline
    $2$ & pedestrian enriched & $6101.1$   &  $0.5399$             & $0.6911$  & $\textbf{0.5586}$      &   $0.7554$\\
    \hline
    \rowcolor[gray]{0.9}
    $3$ & corner case enriched & $6215.7$    &$\textbf{0.5683}$    & $\textbf{0.7173}$  & $0.5384$      &   $0.7517$\\
    \hline
\end{tabular}
}
\label{tab:results}
\end{center}
\end{table*}

\subsection{Human-In-The-Loop Approaches}
As we are in the third wave of AI systems, there has been increased interest in human-in-the-loop (HITL) and machine learning approaches, where humans interact with machines to combine human and machine intelligence to solve a given problem~\cite{Wu2021hitlsurvey}. For this purpose simulators were used to improve AI systems by means of human experience or to study human behavior in field trials, which will now be briefly summarized. 

The use of simulators and human drivers is applied in reinforcement learning, where an agent learns faster from human experience. For instance, the authors of~\cite{wu2021humanintheloop} propose a real-time Deep Reinforcement Learning (Hug-DRL) method based on human guidance, where a person can intervene in driving situations when the agent makes mistakes. These driving errors can be fed directly back into the agents' training procedure and improve the training performance significantly. Furthermore, a method for generating decision corner cases for connected and automated vehicles (CAVs) for testing and evaluation purposes is proposed in~\cite{Sun2021}. For this, the behavioral policy of background vehicles (BV) is learned through reinforcement learning and Markov's decision process, which leads to a more aggressive interaction with the CAV which forces more corner cases under test conditions. The tests take place on the highway and include lane changes or rear-end collisions.

Human behaviors can also be analyzed using driving simulators. For example, in~\cite{Driggs-Campbell2015} a realistic test rig including a steering wheel and pedals for data collection was developed. Therefore, thirteen subjects were recruited to drive on different routes while being distracted by static or dynamic objects or by answering messages on their cell phones. By adding nonlinear human behaviors and using realistic driving data, the authors have been able to predict human driving behavior more accurately in testing. Another driving simulator was presented in~\cite{gomez2018simulator} to develop and evaluate safety and emergency systems. The control units are connected to a generic simulator for academic robotics which uses the Modular Open Robots Simulation Engine \textit{MORSE}~\cite{Echeverria2011morse}. They used an experiment with four road users, one human driver and three vehicles driving in pilot mode and forcing two out of $36$ collision situations (\textit{a lead vehicle stopped} and \textit{a vehicle changing lanes}) defined by the National Traffic Safety Administration (NHTSA). The impact of a driver assistance system on the driver was one of the factors studied.

\begin{figure*}[htp]
\begin{subfigure}{.5\textwidth}
    \begin{subfigure}{.48\textwidth}
        \includegraphics[width=\linewidth]{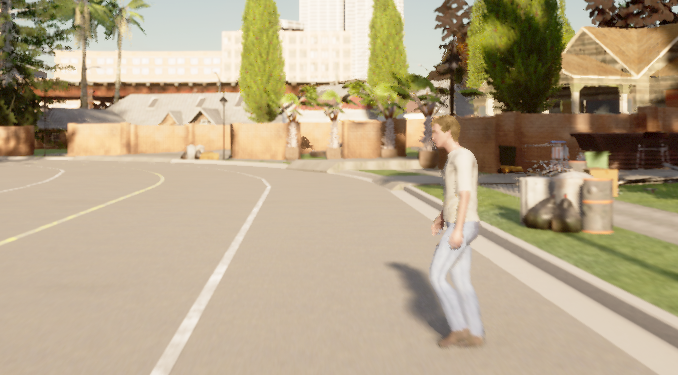}~
    \end{subfigure}
    \begin{subfigure}{.48\textwidth}
        \includegraphics[width=\linewidth]{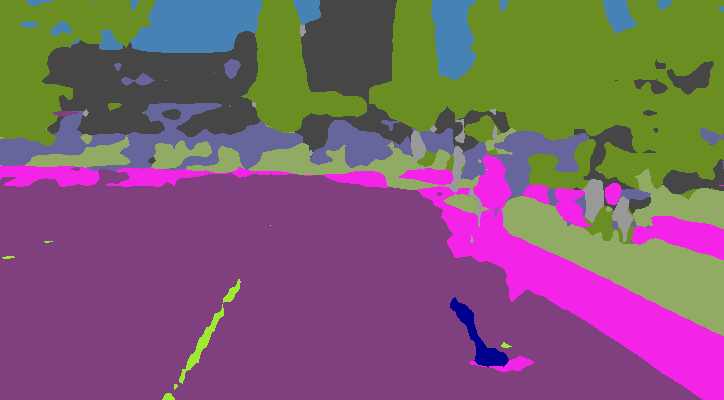}~
    \end{subfigure}
    \hfill
\end{subfigure}
\begin{subfigure}{0.5\textwidth}
    \begin{subfigure}{.48\textwidth}
        \includegraphics[width=\linewidth]{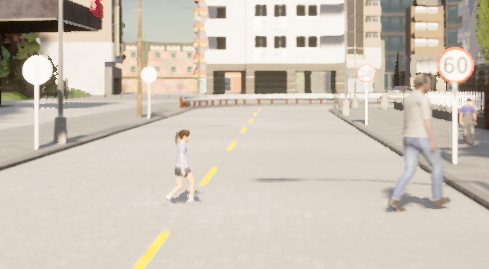}~
    \end{subfigure}
    \begin{subfigure}{.48\textwidth}
        \includegraphics[width=\linewidth]{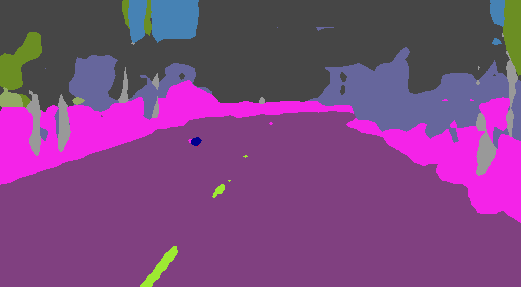}~
    \end{subfigure}
    \hfill
    
\end{subfigure}
\caption{Two examples of a corner case where the safety driver had to intervene to avoid a collision due to the poor prediction of the semantic segmentation network.}
\label{fig:example_cc}
\end{figure*}

\section{Experimental Setup} \label{sec:setup}
\subsection{Driving Simulator}
Targeted enrichment of training data with safety-critical driving situations is essential to increase the performance of AI algorithms. Since the generation of corner cases in the real world is not an option for safety reasons, generation remains in the synthetic world, where specific critical driving situations can be simulated and recorded. For this purpose, the autonomous driving simulator CARLA~\cite{dosovitskiy2017carla} is used. It is open-source software for data generation and/or testing of AI algorithms. It provides various sensors to describe the scenes such as camera, LiDAR and RADAR and delivers ground truth data. 
CARLA is based on the Unreal Engine game engine~\cite{unrealengine}, which calculates and displays the behavior of various road users while taking physics into account, thus enabling realistic driving. Furthermore, the world of CARLA can be modified and adapted to one's own use case with the help of a Python API.

For our work, we used the API to modify the script for manual control from the CARLA repository. In doing so, we added another sensor, the inference sensor, which evaluates the CARLA RGB images in real-time and outputs the neural network semantic prediction on the screen. An example is shown in \Cref{fig:carla_view}. By connecting a control unit including a steering wheel, pedals and a screen, to CARLA, we make it possible to control a vehicle with 'the eyes of the AI' in the synthetic world of CARLA. We also connected a second control unit with the same components to the simulator, so that it is possible to control the same vehicle with 2 different control units, see \Cref{fig:test_rig}. The second control unit is thus operated on the basis of CARLA clear image and can intervene at any time. It always has priority and triggers that the past $3$ seconds of driving, which are buffered, are written to the dataset on the hard disk. In order for the semantic driver to follow the traffic rules in CARLA, the script had to be modified additionally. The code has been modified to display the current traffic light phase in the upper right corner and the speed in the upper center.

\subsection{Test Rig}
The test rig consists of the following components: a workstation with CPU, 3x GPU Quadro RTX 8000, $2$ driving seats, $2$ control units (steering wheel with pedals), one monitor for each control unit and two monitors for the control center. The driving simulator software used is the open-source software CARLA version $0.9.10$. 

\subsection{Dataset for Initial Training and Testing}
For training, a custom dataset was generated using CARLA 0.9.10, consisting of $85$ scenes with $60$ frames each. In addition, there is a validation dataset with 20 scenes. The dataset was generated on seven maps with one fps and contains the corresponding semantic segmentation image in addition to the rendered synthetic image. The maps include the five standard maps in CARLA and two additional maps that offer a mix of city, highway and rural driving. Various parameters can be set in CARLA we focused on the number of Non-Player-Characters (NPCs), including cars, motorcycles, bicycles and pedestrians, and on environment parameters such as sun position, wind and clouds. Depending on the size of the map, the number of NPCs ranged from $50$ to $150$.

The clouds and wind parameters can be set in the range between $0$ and $100$, with $100$ being the highest value. The wind parameter is responsible for the movement of tree limbs and passing clouds and was in the range of $0$ and $50$. The cloud parameter describes the cloudiness, where $0$ means that there are no clouds at all and $100$ that the sky is completely covered with clouds. We have chosen values between $0$ and $30$. The altitude describes the angle of the sun in relation to the horizon of the CARLA world, with values between $-90$ (midnight) and $90$ (midday). Values between $20$ and $90$ were used for our purpose. The other environmental parameters like rain, wetness, puddles or fog are set to zero. The parameters are chosen so that the scenes reflect everyday situations with a natural scattering of NPCs and in similar good weather. During data generation, the movement of all NPCs was controlled by CARLA.

Furthermore, $21$ corner case scenes were used as test data, each containing $30$ frames. Another test dataset containing $21$ standard scenes without corner cases serves as a comparison, each containing $30$ frames.

\begin{table*}[ht]
\begin{center}
\caption{Corner case appearances on Fast-SCNN trained with 3 different datasets.}
\scalebox{0.8}{
\begin{tabular}{|c|l||c|c|c|a|c|a|c|}
	\hline
	\multicolumn{2}{|c||}{\textbf{dataset}} & \textbf{driven distance $d$}& \textbf{driven time $t$} & \textbf{number CCes} & 
	\textbf{mean}$_{d_{CC}}$ & \textbf{std}$_{d_{CC}}$ & \textbf{mean}$_{t_{CC}}$ & \textbf{std}$_{t_{CC}}$  \\
	\hline
	\textit{no.}&				  & $[km]$ & $[min]$ & [-] 	& $[km/CC]$ & $[km/CC]$	& $[min/CC]$	& $[min/CC]$ \\
	\hline
	$1$ & natural distribution    &  $121.32$ & $411$ & $13$  	&  $7.73$ & $14.25$ & $25.93$ &	$39.60$ \\
	\hline
	$2$ & pedestrian enriched     &  $163.09$ & $500$  & $21$ &	$7.52$ &$10.47$	& $23.25$ &$28.72$ \\
	\hline
	$3$ & corner case enriched    &	$153.38$ &  $528$  & $11$  & $\textbf{13.84}$ & $8.68$	& $\textbf{47.47}$& $31.87$ \\
	\hline
\end{tabular}
}
\label{tab:2ndcampaign}
\end{center}
\end{table*}

\subsection{Training}\label{sec:training}
To drive on the predicted semantic mask, a real-time capable network architecture is needed. For these purposes, the Fast Segmentation Convolutional Neural Network (Fast-SCNN) model was used~\cite{Poudel2019fastSCNN}. It uses two branches to combine spatial details at high resolution and deep feature extraction at lower resolution achieving a mean Intersection over Union (mIoU) of $0.68$ at $123.5$ fps on the Cityscapes dataset~\cite{Cordts2016Cityscapes}. The network was implemented in the python package PyTorch~\cite{Pytorch2019} and training was done on a NVIDIA Quadro RTX $8000$ graphics card.
Sixteen of the $23$ classes available in CARLA were used for training. Cross entropy was used as the loss function and ADAM as the optimization algorithm. A polynomial decay scheduler was also used to gradually reduce the learning rate. 

We intentionally stopped the training after $5$ epochs to increase the frequency of perception errors for the network. The resulting network is sufficiently well trained to recognize the road and all road users, although objects further away are poorly recognized. An example is shown at the top of \Cref{fig:carla_view}.

\subsection{Experimental Design}\label{sec:design}
Two operators conducted the driving campaign and the duration of driving as a safety driver or semantic driver was set at 50:50. Both participants had time to familiarize themselves with the hardware and the CARLA world before the start of the first driving campaign so that driving errors could be minimized. Two driving campaigns are planned; the first campaign will generate targeted corner cases and the second will test whether adding the corner cases included in campaign 1 leads to an improvement in the perception of safety-critical situations. 

\section{\uppercase{Retrieval of Corner Cases}} \label{sec:data_collection}
For the generation of corner cases, we consider the following experimental setup. Two test operators record scenes in our specially constructed test rig (see \Cref{fig:test_rig}), where one subject (safety driver) gets to see the original virtual image and the other (semantic driver) the output of the semantic segmentation network (see \Cref{fig:carla_view}). The test rig is equipped with controls such as steering wheels, pedals and car seats and connected to CARLA to simulate realistic traffic participation.

The corner cases were generated as shown in \Cref{fig:flowchart}. For this purpose, we used a real-time semantic segmentation network from \Cref{sec:training} where visual perception was limited. We note that autonomous vehicles according to~\cite{Favaro2017} $67619.81$ km drive until an accident happens. Using a poorly trained network as a part of our accelerated testing strategy, we were able to generate corner cases after $3.34$ km in average between interventions of the safety driver. We note however that the efficiency of the corner cases was evaluated using a fully trained network. \Cref{fig:example_cc} shows two safety-critical corner cases where the safety driver had to intervene to prevent a collision.

If the safety driver triggered the recording of a corner case the test operators label the corner case with one of four options available (overlooking a walker or a vehicle, disregarding traffic rules, intervening out of boredom) and may leave a comment. Furthermore, the kilometers driven and the duration of the ride are notated. The operators were told to obey the traffic rules and not to drive faster than $50$ km/h during the test drives. After a certain familiarization period, driving errors decreased and sudden braking by the semantic driver was also reduced. The reason for this is that the network partially represents areas as vehicles or pedestrians with fewer pixels. Over time, a learning effect occurred in the drivers to hide such situations because experience showed that there was no object there due to the previous frames.

The rides are tracked and by the intervention of the safety driver the last 3 seconds of the scene are saved. Subsequently, the scenes can be loaded and images saved from the ego vehicle's perspective using the camera and the semantic segmentation sensor. We collect $50$ corner cases before retraining from scratch with a mixture of original and corner case images. For each corner case, the last $3$ seconds are saved at $10$ fps before the intervention by the safety driver. In total, we get $1500$ new frames. When using this corner case data for retraining, we delete the same number of frames from the original training dataset. 

We selected $50$ corner cases in connection with pedestrians. Therefore, the inclusion of corner case scenes into the training dataset significantly increases the average number of pixels with the pedestrian class in the training data. To establish a fair comparison of the efficiency of corner cases as compared to a simple upsampling of the pedestrian class, we created a third dataset that contains approximately the same number of pixels per scene for the pedestrian class as the dataset with the corner cases, see \Cref{tab:results}. 

\begin{figure*}[t]
    \centering
    \captionsetup[subfigure]{labelformat=empty}
    \subfloat[corner case enriched]{\includegraphics[width=0.24\textwidth]{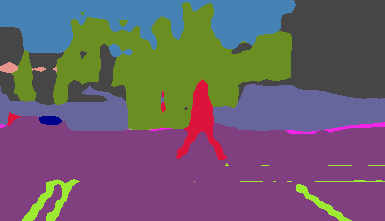}}~
    \subfloat[natural disribution]{\includegraphics[width=0.24\textwidth]{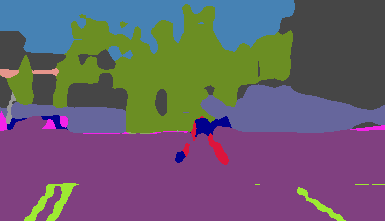}}~
    \subfloat[pedestrian enriched]{\includegraphics[width=0.24\textwidth]{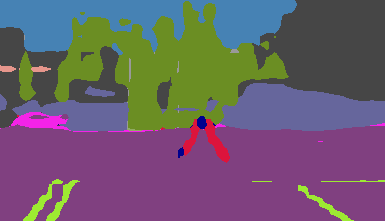}}~
    \subfloat[ground truth]{\includegraphics[width=0.24\textwidth]{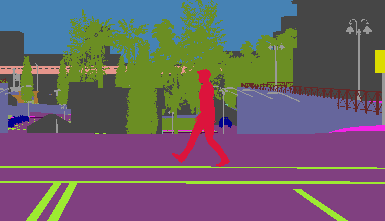}}
    \vspace{5pt}
    \caption{Evaluation on corner case test data shows that the model using corner case data in training recognizes pedestrians better than the model trained with the natural distributed dataset or the dataset which contains more pedestrians.}
    \label{fig:example}
\end{figure*}
\section{\uppercase{Evaluation and Results}} \label{sec:eval}
All results in this section are averaged over $5$ experiments to obtain a better statistical validity. For testing purposes, we generated $21$ additional corner cases for validation. With the same setup as before, we train the Fast-SCNN for $200$ epochs on all three datasets and thereby obtain three networks. \Cref{tab:results} shows the evaluation of all three models on the class pedestrian for the $21$ safety-critical test corner cases. We see that adding corner cases to the training data leads to an improvement in pedestrian detection in safety-critical situations, which can also be shown by an example in \Cref{fig:example}. There we see a situation with a pedestrian crossing the road, with a slope directly behind him that seems to end the road at the level of the horizon. Therefore, the networks that did not have corner cases in the training data seem to have problems with this situation, while the model with corner cases detects the humans much better.

While training the network using naive upsampling of pedestrians does not have any positive effect on the classe's Intersection over Union (IoU) as compared with the original training data, we achieve a gain in the IoU by $2.19\%$ when using the dataset containing corner cases. In addition, the 3 models were tested on a dataset with a natural distribution of pedestrians. Here it can be seen that the model trained with corner cases does not perform as well as the model with the same number of pedestrians. It follows that the model performs better in critical situations, while the models without corner cases perform less well. 

Since our method for generating corner cases in safety-critical situations provides an improvement in detecting pedestrians in safety-critical situations, we launched a second campaign to verify how long it takes driving with the 3 trained networks to generate a corner case. When conducting the second driving campaign, the same operating parameters were set as in the first campaign and the duration until a corner case occurred was recorded. This includes the same maps and weather conditions as well as the same two drivers. However, the two drivers were not aware of the network's underlying training data. \Cref{tab:2ndcampaign} shows the duration and kilometers driven for the different data sets, as well as the occurrence of corner cases during these rides. We see that adding corner cases during training also reduces the frequency until new corner cases reappear.

We therefore demonstrated the benefits of our method to generate corner cases, especially for safety-critical situations. We were also able to show that adding safety-critical corner cases recorded by intentional perceptional distortions improves performance, so future datasets should include such situations.

\section{\uppercase{Conclusion}} \label{sec:conclusion}
Due to the lack of explanation and transparency in the decision-making of today's AI algorithms, we developed an experimental setup that allows to visualize these decisions and thus to allow a human driver to evaluate the driving situations while driving with the eyes of AI, and from this to extract data that includes safety-critical driving situations. Our self-developed test rig provides two human drivers controlling the ego vehicle in the virtual world of CARLA. The semantic driver receives the output of a semantic segmentation network in real-time, based on which she or he is supposed to navigate in the virtual world. The second driver takes the role of the driving instructor and intervenes in dangerous driving situations caused by misjudgments of the AI. We consider driver interventions by the safety driver as safety-critical corner cases which subsequently replaced part of the initial training data. We were able to show that targeted data enrichment with corner cases created with limited perception leads to improved pedestrian detection in critical situations. 

In addition, we take up the idea of the HAI framework and continue the further development of AI by means of human risk perception to identify situations that are particularly important to humans, and thus train the AI precisely where it is particularly challenged by a human perspective.

Future research projects include the use of networks of different quality, changing the weather parameters and provoking accident scenarios so that the number of corner cases can be artificially increased in test operation. In addition, multi-screen driving should be expanded to increase the field of view (FOV) for a more realistic driving experience.
The intervention of the safety driver in the driving situation will also be observed. To this end, criteria for measuring human-machine interaction (HMI) will be developed to track, for example, latency, attention, and intervention due to boredom of the drivers. In addition, contextual and personal factors of human drivers will be investigated to assess uncertainty or anxiety while driving.

\section*{\uppercase{Acknowledgements}}

\noindent The research leading to these results is funded by the German Federal Ministry for Economic Affairs and Climate Action within the project KI Data Tooling under the grant number 19A20001O. We thank Matthias Rottmann for his productive support, Natalie Grabowsky and Ben Hamscher for driving the streets of CARLA and Meike Osinski for the support in the field of human-centered AI.

\bibliographystyle{apalike}
{\small
\bibliography{citation}}

\end{document}